\documentclass[journal]{IEEEtran}

\usepackage{cite}

\ifCLASSINFOpdf
\else
\fi
\usepackage{amsmath,amssymb,amsfonts}
\usepackage[caption=false]{subfig}
\usepackage{threeparttable}
\usepackage{multirow}
\usepackage{cite}
\usepackage{microtype}
\usepackage{dblfloatfix}
\usepackage{tikz}

\usepackage{pifont}
\let\oldding\ding
\renewcommand{\ding}[2][1]{\scalebox{#1}{\oldding{#2}}}
\newcommand{\ineq}[1]{\footnotesize$#1$\normalsize}{}

\newcommand{\mr}[1]{\textcolor{black}{#1}}
\newcommand{\prior}{\text{{SpiNeMap}}}{}

\begin{document}
\bstctlcite{IEEEexample:BSTcontrol}
%
\title{Enabling Resource-Aware Compilation of Spiking Neural Networks to Neuromorphic Hardware}
\title{Enabling Resource-Aware Mapping of Spiking Neural Networks via Spatial Decomposition}
%
%
%

\author{Adarsha~Balaji,
        Shihao~Song,
        Anup~Das,
        Jeffrey~Krichmar,
        Nikil~Dutt,\\
        James~Shackleford,
        Nagarajan~Kandasamy,
        and~Francky~Catthoor
\thanks{A. Balaji, S. Song, A. Das, J. Shackleford, and N. Kandasamy are with the Department
of Electrical and Computer Engineering, Drexel University, Philadelphia,
PA, 19104 USA e-mail: \{shihao.song,anup.das,jas64,nk78\}@drexel.edu}
\thanks{J. Krichmar and N. Dutt are with the Department
of Computer Science, University of California, Irvine, CA, USA.}
\thanks{F. Catthoor is with Imec, Belgium and KU Leuven, Belgium.}
\thanks{Manuscript received Month Day, Year; revised Month Day, Year.}}

\markboth{IEEE Embedded Systems Letters,~Vol.~XX, No.~X, Month~Year}%
{Song \MakeLowercase{\textit{et al.}}: Enabling Resource-Aware Compilation of Spiking Neural Networks to Neuromorphic Hardware}


\maketitle

\begin{abstract}
With growing model complexity, mapping Spiking Neural Network (SNN)-based applications to tile-based neuromorphic hardware is becoming increasingly challenging. This is because the synaptic storage resources on a tile, viz. a crossbar, can accommodate only a fixed number of pre-synaptic connections per post-synaptic neuron. 
For complex SNN models 
that have
many pre-synaptic connections per neuron, some 
connections may need to be pruned after training to fit onto the tile resources, leading to a loss in model quality, e.g., accuracy. 
In this work, we propose a novel unrolling technique that decomposes a neuron function with many pre-synaptic connections into a sequence of homogeneous neural units, where each neural unit is a function computation node, with two pre-synaptic connections. 
\mr{
This spatial decomposition technique significantly improves crossbar utilization and retains all pre-synaptic connections, resulting in no loss of the model quality derived from connection pruning. 
}
We integrate the proposed technique within an existing SNN mapping framework and evaluate it using machine learning applications on the DYNAP-SE  state-of-the-art neuromorphic hardware. 
Our results demonstrate an average 60\% lower crossbar requirement, 9x higher synapse utilization, 62\% lower wasted energy on the hardware, and between 0.8\% and 4.6\% increase in model quality.
\end{abstract}


%
\IEEEpeerreviewmaketitle

\section{Introduction}\label{sec:introduction}
\begin{figure}[h!]
	\centering
	\centerline{\includegraphics[width=0.79\columnwidth]{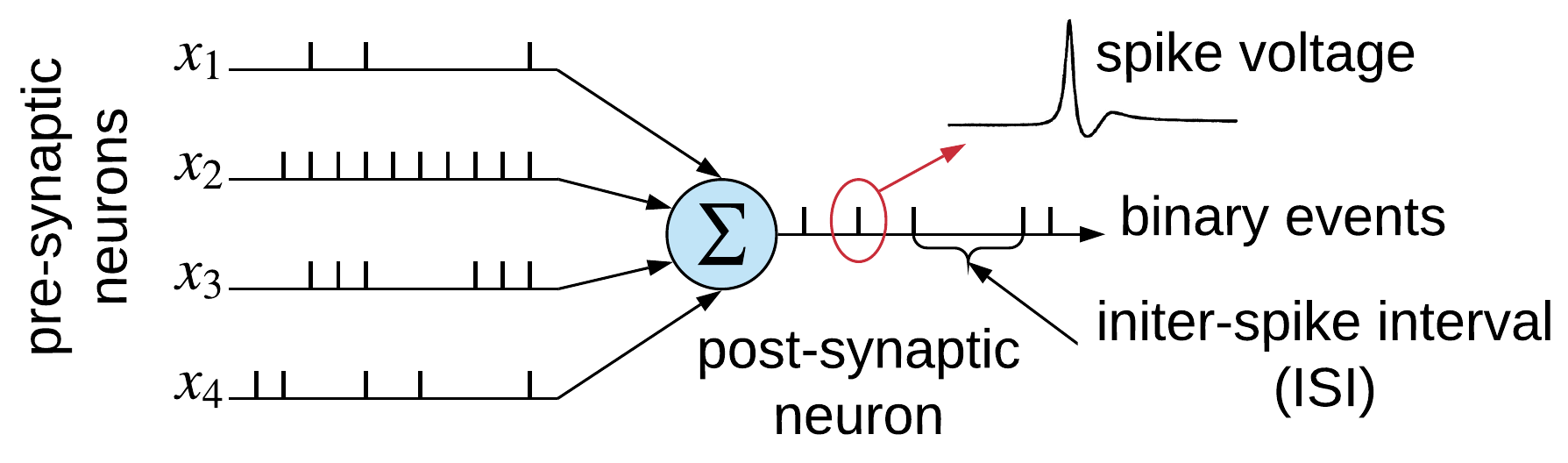}}
	\caption{An example of spiking neural network.}
	\label{fig:snn}
\end{figure}

\IEEEPARstart{S}{piking} Neural Networks (SNNs) are machine learning approaches designed using spike-based computation and bio-inspired learning algorithms~\cite{maass1997networks}. In an SNN, pre-synaptic neurons communicate information encoded in spike train to post-synaptic neurons, via the synapses (see Fig.~\ref{fig:snn}). 
\mr{
One of the key quality metrics in SNNs is the inter-spike interval (ISI) defined as the inverse of the mean firing rate of a neuron, which directly impacts model quality, e.g., accuracy.
}

\begin{figure}[h!]
	\centering
	\centerline{\includegraphics[width=0.79\columnwidth]{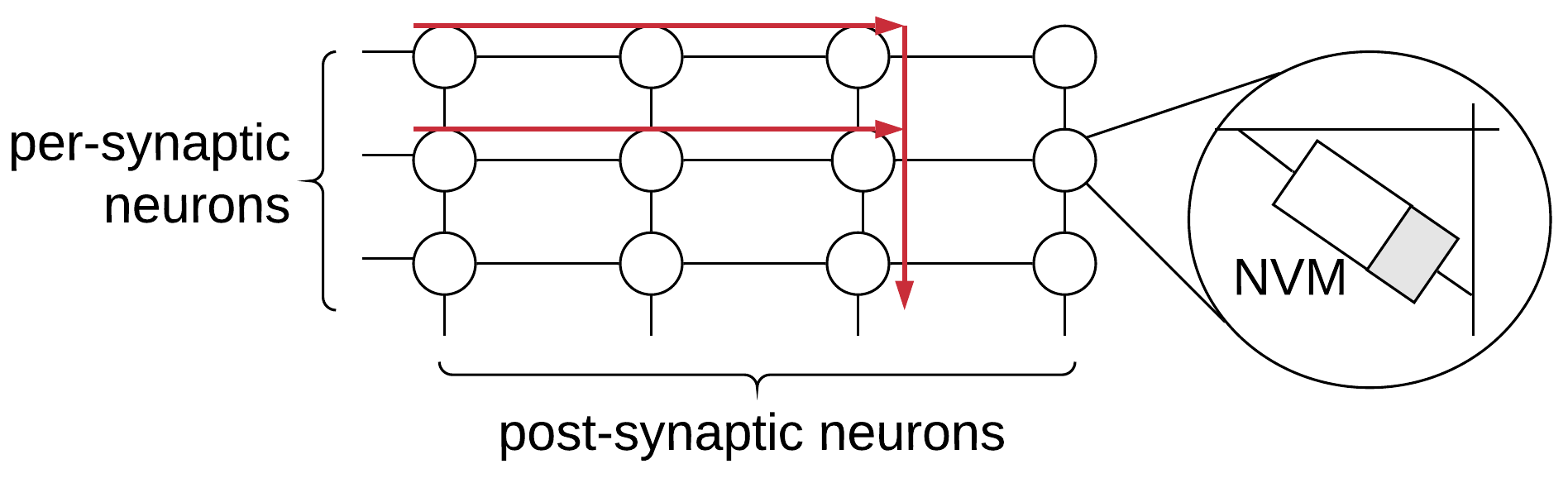}}
	\caption{A crossbar in a neuromorphic hardware.}
	\label{fig:neuromorphic}
\end{figure}

SNNs are executed on tile-based neuromorphic hardware such as DYNAP-SE~\cite{dynapse}. A tile consists of a crossbar~\cite{catthoor2018very}, which is illustrated in Fig.~\ref{fig:neuromorphic}. At each crosspoint of a crossbar there is a synaptic element, which can be implemented using Non-Volatile Memory (NVM)~\cite{Mallik2017}. 
Beside neuromorphic computing, NVMs such as Phase-Change Memory (PCM) are also used as main memory in conventional computers to improve performance and energy efficiency~\cite{palp,mneme,datacon,hebe}.

\mr{
Within each crossbar, 
synaptic weights between pre- and post-synaptic neurons are programmed as conductance of NVM cells.
A pre-synaptic neuron's voltage (applied on a row) is multiplied by the conductance to generate current (according to Ohm's Law).
Current summation occurs on each column according to Kirchoff's Current Law, when integrating excitation from its pre-synaptic neurons.
A neuron circuit takes as its input this current and generates at its output a train of spikes. A spike is generated only if the current is higher than a threshold. The spike firing frequency increases with input current, saturating at a frequency determined by the refractory period of the neuron.
}

\begin{figure}[h!]
	\centering
	\centerline{\includegraphics[width=0.99\columnwidth]{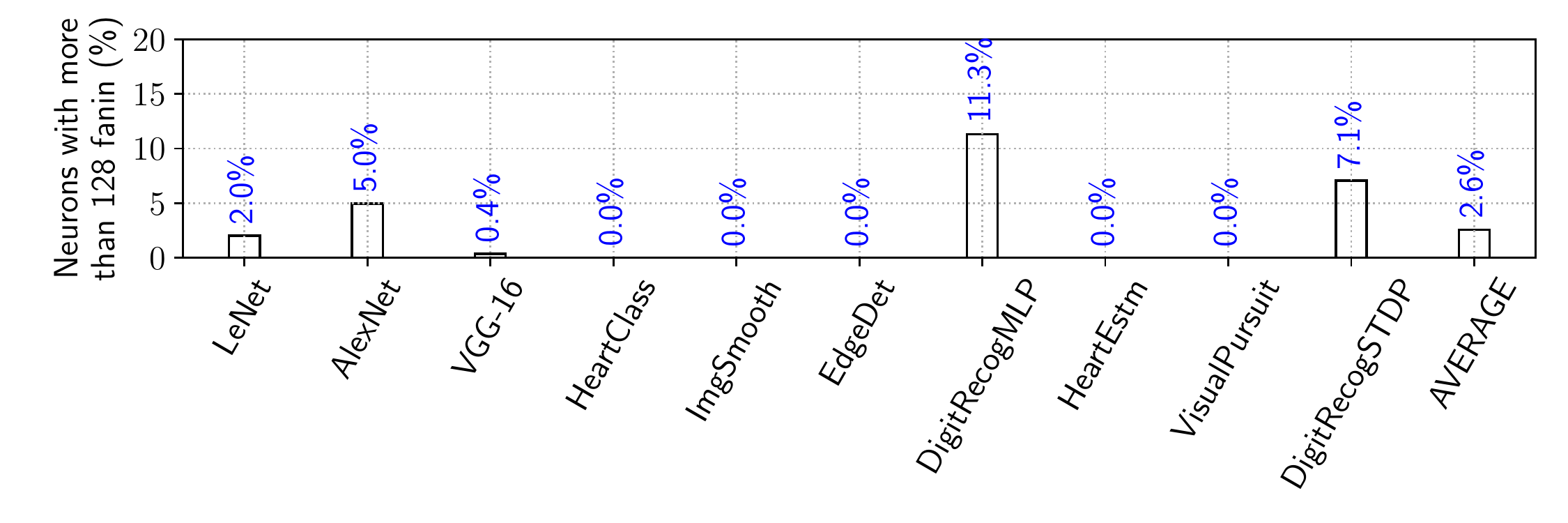}}
	\caption{Fraction of neurons with 128 or more pre-synaptic connections (fanin).}
	\label{fig:fanin}
\end{figure}

A \ineq{n\times n} crossbar in a tile can accommodate only \ineq{n} pre-synaptic connections per post-synaptic neuron (\ineq{n = 128} for the crossbars in DYNAP-SE). 
\mr{Fig.~\ref{fig:fanin} reports the number of neurons with more than 128 pre-synaptic connections, i.e., fanins as a fraction of the total number of neurons in a few standard machine learning models, which are pruned iteratively to eliminate the near-zero weights without loss in accuracy~\cite{han2015deep} (see our design framework in Fig.~\ref{fig:approach}).} 

\mr{We observe that, even after model pruning, on average {2.6\%} of neurons in these models cannot be mapped to the crossbars in DYNAP-SE.} There are two currently-used solutions to this problem -- 1) implement larger crossbars, which increases the power consumption exponentially, and 2) remove synaptic connections, which reduces the model quality.

Instead, we propose an unrolling technique which decomposes each neuron having many pre-synaptic inputs or \textit{fanin} into a sequence of homogeneous neural units, where each neural unit is a function computation node having a maximum fanin-of-two (or FIT). Our technique ensures information integrity as well as the quality of these unrolled machine learning models. Furthermore, the unrolling technique allows denser packing of the homogeneous neural units per crossbar, significantly improving the crossbar utilization. We evaluate our technique using standard machine learning applications and demonstrate an average 60\% lower resource usage and 62\% lower wasted energy. The proposed spatial decomposition technique retains all pre-synaptic connections for each neuron in an SNN, improving model quality between 0.8\% and 4.6\% compared to an existing SNN mapping approach.

\section{Proposed Technique}\label{sec:proposed_approach}
\begin{figure}[h!]
	\centering
	\centerline{\includegraphics[width=0.99\columnwidth]{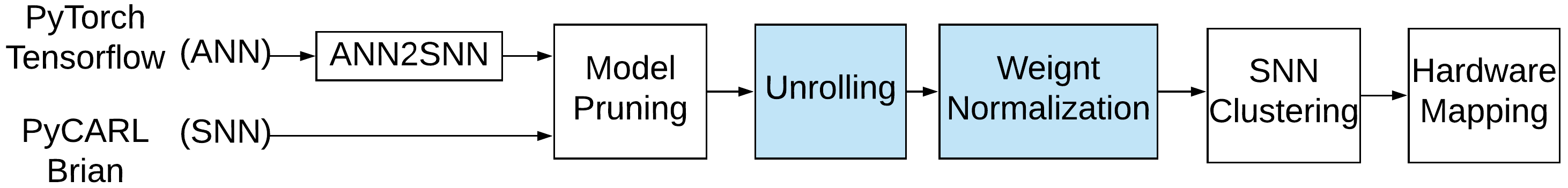}}
	\caption{Integration of the proposed approach in an existing design flow.}
	\label{fig:approach}
\end{figure}

Fig.~\ref{fig:approach} illustrates the integration of the proposed technique (shown in the two colored boxes) inside an existing design flow \prior{}~\cite{spinemap} for mapping SNNs to the hardware.\footnote{\mr{The proposed approach can be integrated with many other mapping approaches such as the data flow-based performance-oriented SNN compilation technique of~\cite{dfsynthesizer,das2018mapping,balaji2019frameworkISVLSI,das2018dataflow}, the circuit aging oriented SNN mapping technique of \cite{reneu,frameworkCAL,NeuromorphicLR}, endurance-aware mapping technique of~\cite{twisha_thermal,twisha_endurance}, and the SNN compiler of \cite{ji2018bridge,rtmJSPS}.}}
This design flow incorporates 1) Artificial Neural Network (ANN) models written in PyTorch~\cite{paszke2019pytorch} and Tensorflow~\cite{abadi2016tensorflow}, and 2) SNN models written in PyCARL~\cite{pycarl} and Brian~\cite{goodman2009brian}. 
For ANN models, analog operations are first converted to spike-based operations using the \texttt{ANN2SNN} converter~\cite{HeartClassJolpe}. 
\mr{
These models are first pruned to eliminate near-zero weights~\cite{han2015deep}. This is to keep the model size small for embedded platforms. The pruned models are then presented to the proposed unrolling technique, which compiles these models to crossbar-based hardware.
}

Our proposed technique works in two steps -- 1) the unrolling step, which decomposes the SNN to limit the number of pre-synaptic connections for each neuron, and 2) the weight normalization step, which ensures the quality of the decomposed model. The latter is then clustered and mapped to the neuromorphic hardware using \prior{}~\cite{spinemap}. 


\subsection{Spatial Decomposition using Model Unrolling}
\begin{figure}[h!]
	\centering
	\centerline{\includegraphics[width=0.99\columnwidth]{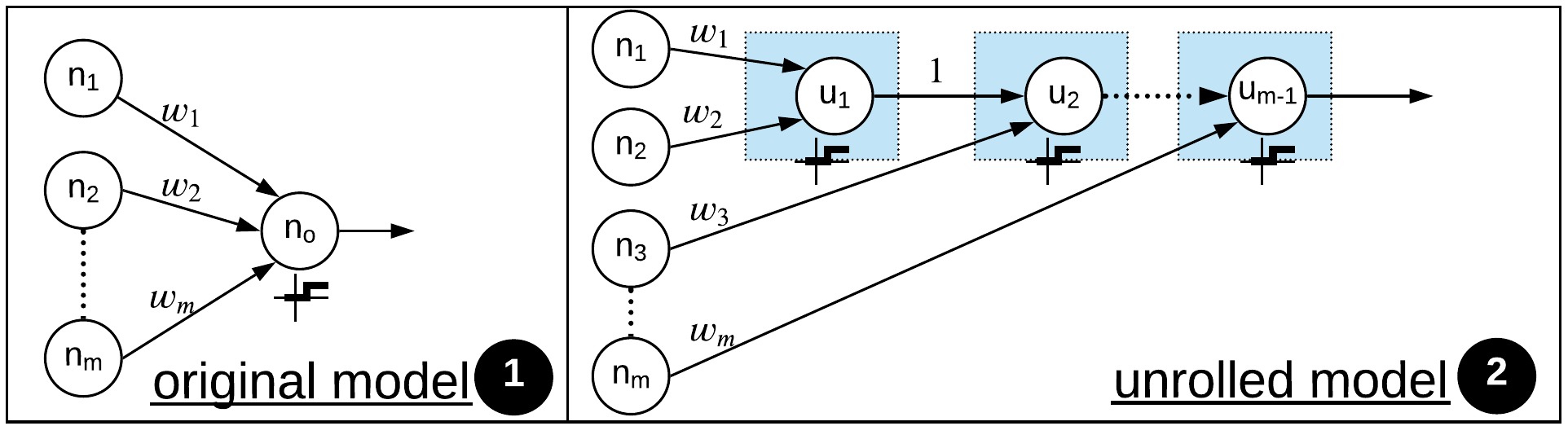}}
	\caption{\mr{Unrolling a neuron functionality.}}
	\label{fig:unrolling}
\end{figure}

We propose an {unrolling} approach, which decomposes a neuron function computation with many fanins into a sequence of homogeneous \textit{neural units}, where each neural unit is a computation node with a maximum fanin-of-two ({FIT}).
Fig.~\ref{fig:unrolling} illustrates the decomposition (\ding{183}) of the neuron function shown in \ding{182}. Here, one \ineq{m}-input neuron function is decomposed into \ineq{(m-1)} two-input neural units connected in sequence.

\mr{
The neuron function \ineq{y_o = f\left(\sum_{i=1}^m n_i\cdot w_i\right)}
is represented as
\begin{equation}
\label{eq:decomposed_nn}
\footnotesize y_o = f(u_{m-1}), \text{ where } u_i = \begin{cases}
        f(n_1\cdot w_1 + n_2\cdot w_2), \text{ for } i = 1
        \\
        f(u_{i-1} + n_{i+1}\cdot w_{i+1}), \text{ otherwise}.
        \end{cases}
\end{equation}
where \ineq{f} represents the neuron functionality of generating spike trains with a mean firing rate proportional to its input excitation, \ineq{n_1,n_2,\cdots,n_m} are the \ineq{m} pre-synaptic neurons of the post-synaptic neuron \ineq{n_o}, and \ineq{w_1,w_2,\cdots,w_m} are the corresponding synaptic weights.
}
The total number of FIT neural units generated from a neural network with \ineq{N} neurons is \ineq{\mathcal{N} = \sum_{i=1}^N(m_i-1)}, where \ineq{m_i} is the fanin of neuron \ineq{n_i}. 

\begin{figure*}[t!]
	\centering
	\centerline{\includegraphics[width=1.99\columnwidth]{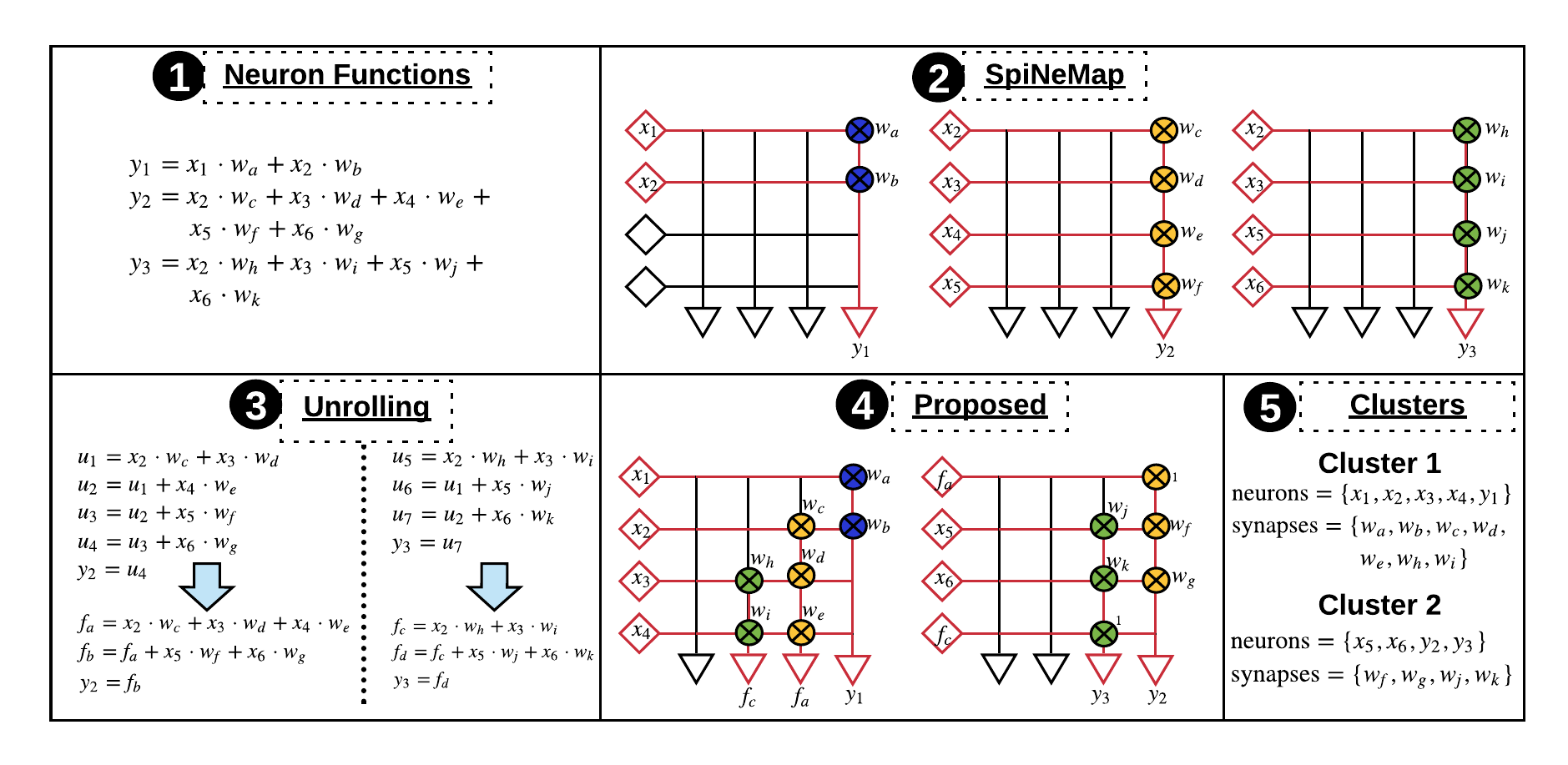}}
	\caption{Demonstration of SNN clustering using \prior{}~\cite{spinemap} (\ding{183}) and the proposed approach (\ding{185}).}
	\label{fig:clustering_demo}
\end{figure*}

\subsection{Weight Normalization}
\mr{
We apply weight normalization to optimize the synaptic weights \ineq{w_i} of the unrolled model (Fig.~\ref{fig:unrolling}\ding{182}) such that the firing rate of a neuron in this model is proportional to its input activation in the original model (Fig.~\ref{fig:unrolling}\ding{183}). The weight normalization is performed for each decomposed neural unit and the weight normalization factor is applied to all its pre-synaptic weights. Using Fig~\ref{fig:unrolling}, the weight updates are
\begin{equation}
    \label{eq:batch}
    \footnotesize w_{i} = \begin{cases}
        w_{i} / S_\text{norm}^1, \text{ for } i = 1,2
        \\
        w_{i} / S_\text{norm}^{i-1}, \text{ otherwise}.
        \end{cases} \text{~and~} w_{u_i,u_j} = 1/S_\text{norm}^j
\end{equation}
The normalization factor is computed as the maximum activation on the corresponding synaptic weight in the original model using a batch from the training set, i.e.,
\begin{equation}
    \label{eq:factor}
    \footnotesize S_\text{norm}^i = \begin{cases}
        k\cdot \max\{a_1+a_2\} \text{ for } i = 1
        \\
        k\cdot\max\{a_{i+1}\} \text{ otherwise}.
        \end{cases},
\end{equation}
where \ineq{a_i} is the activation on the synaptic weight \ineq{w_i} in the original model and the scaling factor \ineq{k} is used to limit the mean firing rate of a neuron to lower energy consumption on the hardware~\cite{HeartClassJolpe}.
}
The weight normalization overhead can be reduced by allowing non-uniform decomposition of neuron functions. This is part of our future exploration.

\subsection{Motivating Example}
Fig.~\ref{fig:clustering_demo} provides a motivating example demonstrating 1) how the proposed technique results in a difference in the SNN mapping to hardware when compared to \prior{} and 2) the improvements in model quality.


This example illustrates the mapping of 
three neuron functions \ineq{y_1, y_2} and \ineq{y_3}, shown in \ding{182} to a neuromorphic hardware consisting of \ineq{4\times 4} crossbars.
\prior{} and other similar techniques will use three crossbars to implement the three functions, as shown in \ding{183}.
Since a crossbar can accommodate only a limited number of pre-synaptic connections per output neuron (4 in this example), the component \ineq{x_6\cdot w_g} of neuron function \ineq{y_2} cannot be mapped to the crossbar. This may result in a degradation of the model quality.

In the proposed technique, 
neuron functions \ineq{y_2}, with fanin of 5 and \ineq{y_3}, with fanin of 4 are decomposed to generate homogeneous neural computation units \ineq{u_1, u_2, \cdots,u_7} as shown in \ding{184}.
The neuron function \ineq{y_1} is not decomposed because the proposed technique unrolls only those neuron functions that have fanin greater than 2. 
\mr{
During packing of these units to a crossbar with a limited number of input ports, some of the decomposed units may need to be combined to generate larger units.
In this example, the new computations are represented using functions \ineq{f_a,f_b,f_c} \& \ineq{f_d}, where \ineq{f_a} and \ineq{f_c} can be implemented on the first crossbar alongside \ineq{y_1}, while \ineq{f_b (y_2)} and \ineq{f_d (y_3)} are implemented on a second cluster (\ding{185}).
}
Finally, the two generated clusters are shown in \ding{186}.

We make the following three key observations. First, the proposed technique only uses two crossbars to implement the three neuron functionalities \ineq{y_1, y_2} and \ineq{y_3}, one less than that used by \prior{}. This reduces hardware requirement and improves energy consumption. Second, it does not eliminate any component of the neuron functionality, which improves the model quality. Third, it increases the crossbar utilization.

\section{Results and Discussion}\label{sec:results}
\subsection{Evaluation Methodology}
We evaluated 10 machine learning applications that are representative of the three most commonly used neural network classes --- convolutional neural network (CNN), multi-layer perceptron (MLP), and recurrent neural network (RNN).
\mr{
Table~\ref{tab:apps} summarizes the
original models 
and their baseline accuracy, which is assessed from the inter-spike interval~\cite{spinemap}. 
}

\begin{table}[h!]
	\renewcommand{\arraystretch}{0.8}
	\setlength{\tabcolsep}{2pt}
	\caption{Applications used to evaluate our approach.}
	\label{tab:apps}
	\centering
	\begin{threeparttable}
	{\fontsize{6}{10}\selectfont
		\begin{tabular}{cc|ccl|c}
			\hline
			\textbf{Class} & \textbf{Applications} & \textbf{Synapses} & \textbf{Neurons} & \textbf{Topology} & \textbf{Accuracy}\\
			\hline
			\multirow{4}{*}{CNN} & LeNet & 282,936 & 20,602 & CNN & 85.1\%\\
			& AlexNet & 38,730,222 & 230,443 & CNN & 90.7\%\\
			& VGG16 & 99,080,704 & 554,059 & CNN & 69.8 \%\\
			& HeartClass~\cite{HeartClassJolpe,das2018heartbeat} & 1,049,249 & 153,730 & CNN & 63.7\%\\
			\hline
			\multirow{3}{*}{MLP} & DigitRecogMLP & 79,400 & 884 & FeedForward (784, 100, 10) & 91.6\%\\
			& EdgeDet \cite{carlsim} & 114,057 &  6,120 & FeedForward (4096, 1024, 1024, 1024) & 100\%\\
			& ImgSmooth \cite{carlsim} & 9,025 & 4,096 & FeedForward (4096, 1024) & 100\%\\
			\hline
 			\multirow{3}{*}{RNN} & HeartEstm \cite{HeartEstmNN} & 66,406 & 166 & Recurrent Reservoir & 100\%\\
 			& VisualPursuit \cite{Kashyap2018} & 163,880 & 205 & Recurrent Reservoir & 47.3\%\\
 			& DigitRecogSTDP \cite{Diehl2015} & 11,442 & 567 & Recurrent Reservoir & 83.6\%\\
			\hline
	\end{tabular}}
	\end{threeparttable}
\end{table}

We evaluated these applications on the DYNAP-SE neuromorphic hardware~\cite{dynapse}, where each tile has one \ineq{128\times 128} crossbar.
Table \ref{tab:hw_parameters} reports the hardware parameters of DYNAP-SE.

\begin{table}[h!]
    \caption{Major simulation parameters extracted from \cite{dynapse}.}
	\label{tab:hw_parameters}
	\centering
	{\fontsize{6}{10}\selectfont
		\begin{tabular}{lp{5cm}}
			\hline
			Neuron technology & 28nm FD-SOI\\
			\hline
			Synapse technology & Hfo2-based OxRRAM\\
			\hline
			Supply voltage & 1.0V\\
			\hline
			Energy per spike & 50pJ at 30Hz spike frequency\\
			\hline
			Energy per routing & 147pJ\\
			\hline
			Switch bandwidth & 1.8G. Events/s\\
			\hline
	\end{tabular}}
\end{table}

\subsection{Hardware Requirement}
\label{sec:crossbar_count}
\begin{figure}[h!]
	\centering
	\centerline{\includegraphics[width=0.99\columnwidth]{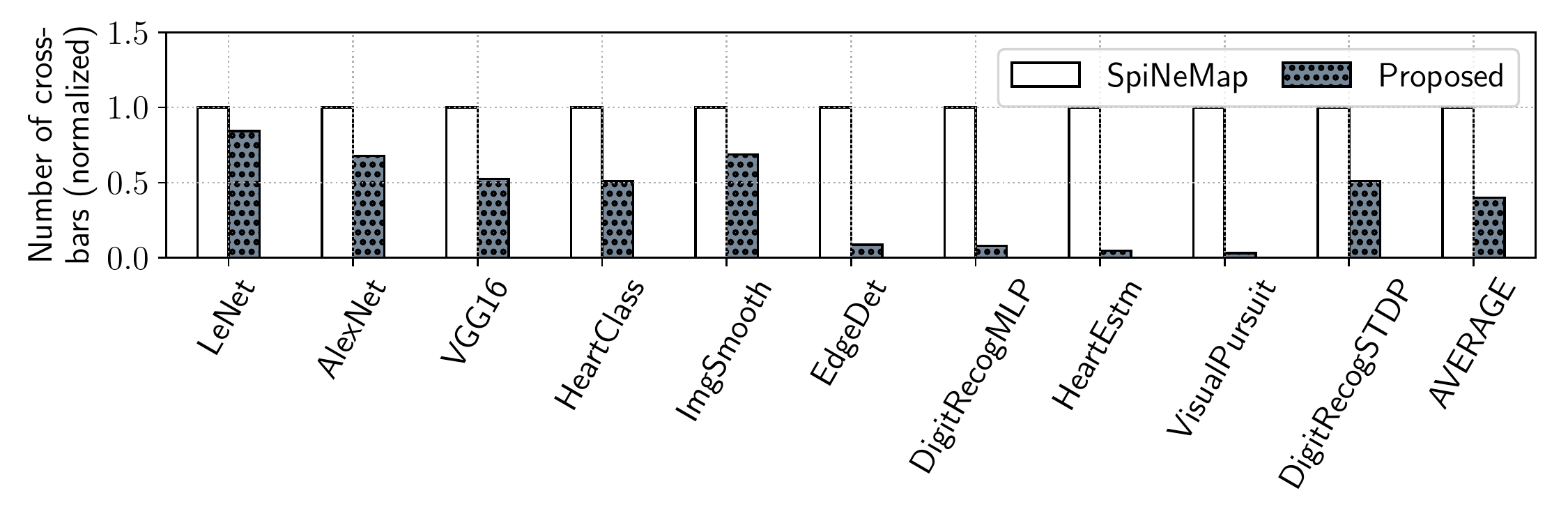}}
	\caption{Crossbars needed for the evaluated applications.}
	\label{fig:cluster_count}
\end{figure}

Figure \ref{fig:cluster_count} compares the number of crossbars required by \prior{} and the proposed technique for each of the evaluated applications. Crossbar numbers are normalized to \prior{}.
We make the following \textit{four} observations.

\textit{First}, on average, the proposed technique requires 60\% fewer crossbars than \prior{} to execute these applications. This reduction is because the unrolling technique allows to densely pack the fanin synapses of different neurons into the same crossbar, reducing the overall crossbar count. This reduces the hardware energy consumption.
\textit{Second}, the number of crossbars required in the proposed technique is 10x lower than \prior{} for EdgeDet, even though EdgeDet has no neurons with more than 128 fanin synapses. This is because many neurons in EdgeDet have fanin close to the maximum limit that a crossbar in DYNAP-SE can accommodate. Therefore, these high-fanin neurons cannot be packed in the same crossbar by \prior{}, needing a separate crossbar for each of these neurons. The proposed technique, on the other hand, unrolls a high-fanin neuron to create a structure with a maximum-fanin-of-two. This results in a denser packing of crossbars and a correspondingly lower crossbar requirement. \textit{Third}, for CNN-based applications (LeNet, AlexNet, VGG16, and HeartClass), the proposed technique requires 37\% fewer crossbars than \prior{}. We observe that the reduction over \prior{} is higher for networks with more layers; the reduction is 16\% for LeNet, compared to 48\% for VGG16. This is because, with more layers, the proposed technique has more freedom to improve crossbar utilization (see Sec. \ref{sec:cluster_utilization}).
\textit{Finally}, for RNN-based applications (HeartEstm, VisualPursuit, and DigitRecogSTDP), the proposed approach requires 80\% fewer crossbars than \prior{}.
This is due to the cyclic nature of connections between neurons in these applications. \prior{} cannot optimally pack neurons in cyclic connections to the crossbars. The proposed technique decomposes a cyclic connection to pack its neurons densely into crossbars, reducing the crossbar requirement.

\begin{figure}[h!]
	\centering
	\centerline{\includegraphics[width=0.99\columnwidth]{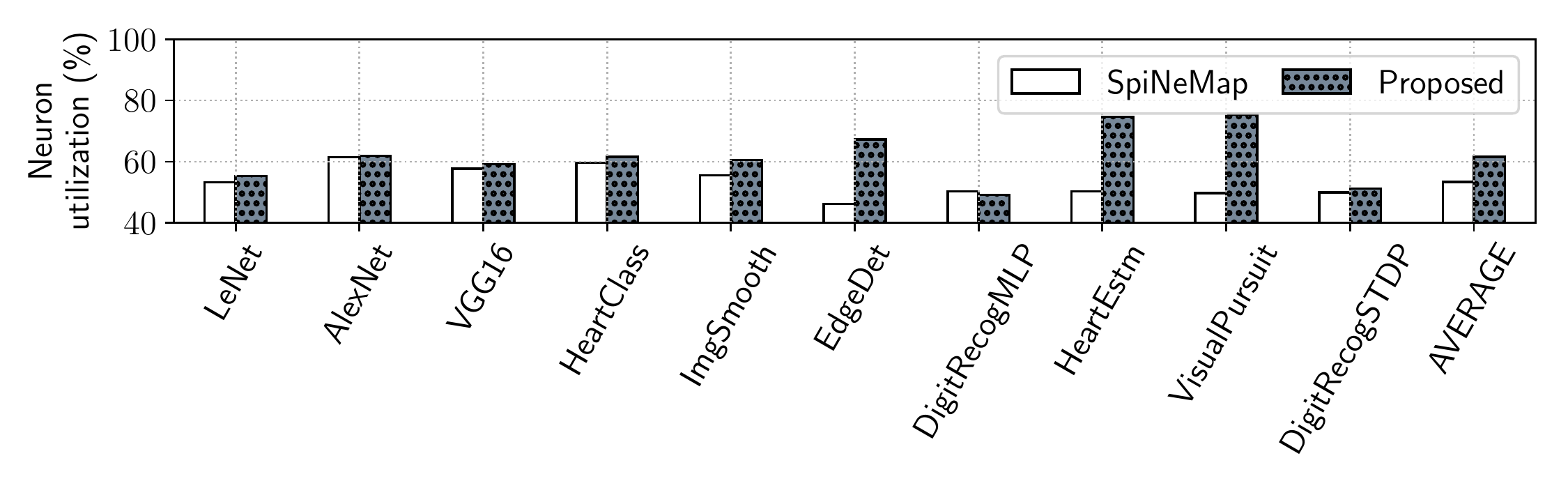}}
	\caption{Neuron utilization.}
	\label{fig:cluster_io}
\end{figure}

\begin{figure}[h!]
	\centering
	\centerline{\includegraphics[width=0.99\columnwidth]{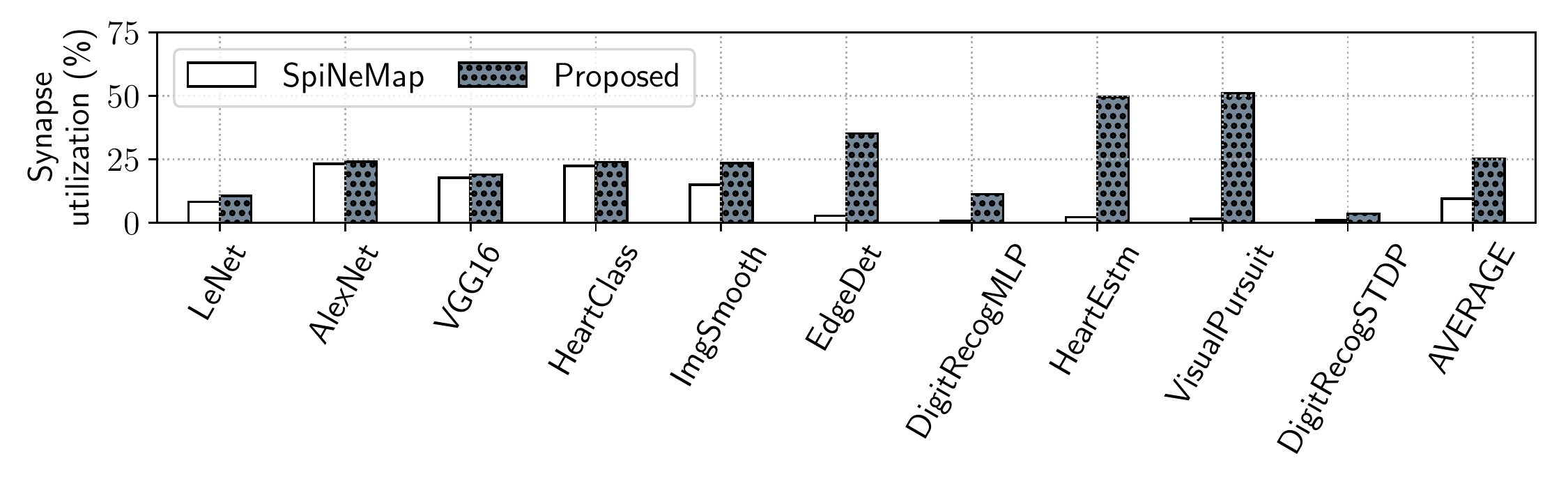}}
	\caption{Synapse utilization.}
	\label{fig:cluster_memory}
\end{figure}

\subsection{Crossbar Utilization}\label{sec:cluster_utilization}
Figures \ref{fig:cluster_io} and \ref{fig:cluster_memory} compare respectively, the neuron and synapse utilization of \prior{} and the proposed technique on DYNAP-SE for each evaluated application. 
We observe that the average neuron utilization of \prior{} is 53\% and of the proposed technique is 62\% for these applications.
The average synapse utilization of \prior{} is only 9\% and of the proposed technique is 25\% for these applications. 
The reason for the high crossbar utilization is because of the proposed spatial decomposition technique, which allows to densely pack fanin and fanout synapses from different neurons in the same crossbar, improving utilization. Higher utilization leads to lower wasted energy as reported next.



\subsection{Wasted Energy}\label{sec:wasted_energy}
\begin{figure}[h!]
	\centering
	\centerline{\includegraphics[width=0.99\columnwidth]{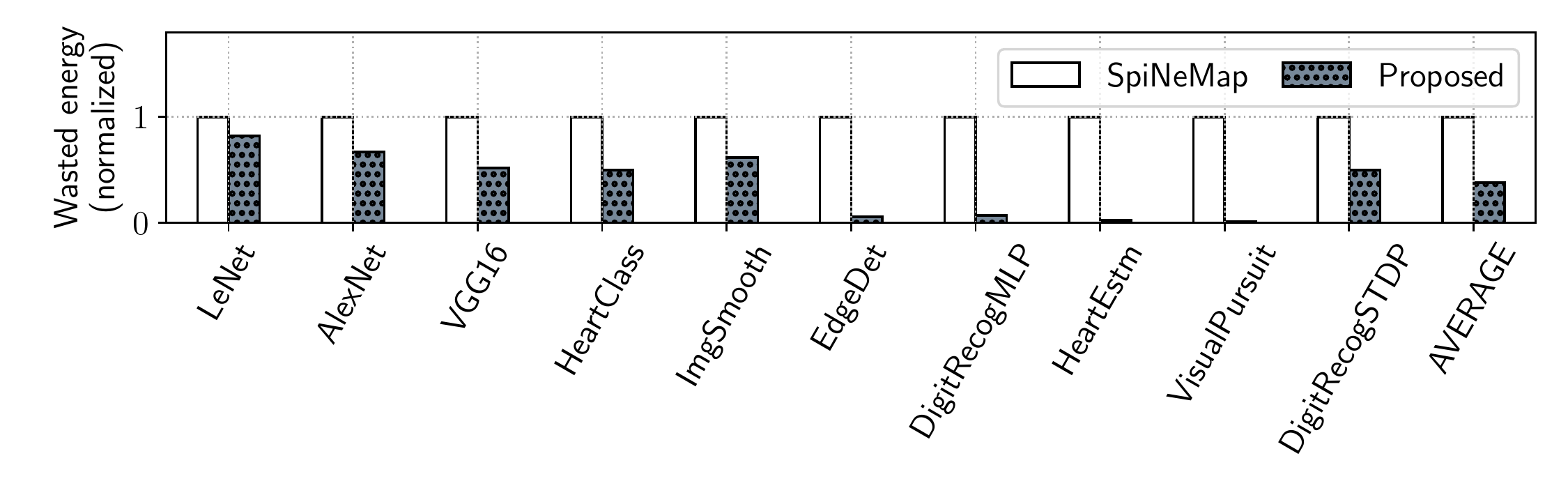}}
	\caption{Wasted energy on DYNAP-SE.}
	\label{fig:wasted_energy}
\end{figure}

Figure \ref{fig:wasted_energy} reports the energy wasted for each evaluated application on DYNAP-SE using the proposed technique, normalized to \prior{}. The wasted energy incorporates the neurons and synapses in each crossbar that are not utilized during the execution of these applications.
We observe that 
the energy wasted using the proposed technique is on average 62\% lower than \prior{}. This significant reduction is due to 1) the reduction in the use of crossbars and 2) the increase in the utilization of neurons and synapses in each crossbar.

\subsection{Model Quality}\label{sec:model_quality}
\begin{figure}[h!]
	\centering
	\centerline{\includegraphics[width=0.99\columnwidth]{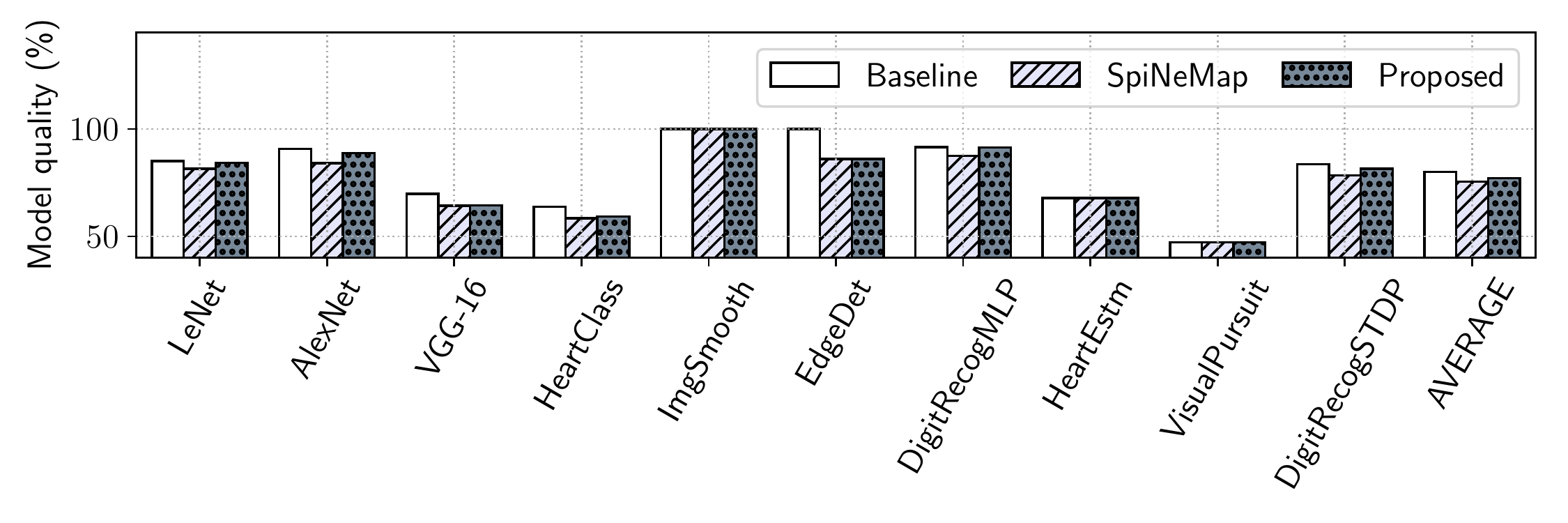}}
	\caption{Model quality for the evaluated applications.}
	\label{fig:quality}
\end{figure}

Figure~\ref{fig:quality} plots the model quality i.e., the accuracy of each evaluated application on DYNAP-SE using \prior{} and the proposed technique. For comparison, the quality using software simulation is also reported in the figure as Baseline. We make the following \textit{two} key observations.

\textit{First}, the accuracy loss for some applications such as ImgSmooth, EdgeDet, HeartEstm, and VisualPursuit are the same for \prior{} and the proposed technique. This is because no neuron in these applications has more than 128 fanin synapses (see Fig.~\ref{fig:fanin}) and therefore, no pre-synaptic connections are eliminated by \prior{}. So, the quality of the two techniques are the same. However, for these applications, the proposed technique is significantly better than \prior{} in terms of crossbar usage, their utilization, and the wasted energy.
Second, for all other applications that do not initially fit on the available crossbar space, the quality of the proposed technique is better than \prior{} by average 3\% (between 0.8\% and 4.6\%). For these applications, the proposed technique is better in terms of all the evaluated metrics.


\section{Discussion}
\mr{
We note that, the decomposition technique proposed here negatively impacts accuracy and energy in following two aspects.
First, by decomposing a large neuron function, the proposed technique maps some of the decomposed synaptic connections on the shared interconnect of a neuromorphic hardware, negatively impacting spike latency. This can lower the model accuracy. However, the SpiNeMap framework, which integrates the decomposition technique minimizes such impact by intelligent cluster mapping and placement on the hardware~\cite{spinemap}. Furthermore, by not eliminating any synaptic connections, the proposed technique, in fact, improves accuracy compared to SpiNeMap (see Sec.~\ref{sec:model_quality}). 
Second, with additional synapses mapped to the shared interconnect, the energy consumption on the interconnect increases. However, we find this increase in energy is much lower than the energy savings obtained by reducing the crossbar usage (see Sec.~\ref{sec:wasted_energy}).
}


\section{Conclusion}\label{sec:conclusions}
We present a technique to map SNN-based applications to crossbar-based neuromorphic hardware. The proposed technique involves unrolling of neurons, which decomposes a complex neuron functionality into a sequence of homogeneous neural units, where each neural unit is a fanin-of-two (FIT) neuron.
\mr{
The unrolling technique significantly improves crossbar utilization and ensures information integrity, resulting in no loss of model quality derived from connection pruning.
}
We integrate this unrolling technique inside an existing SNN mapping framework and evaluate it using machine learning applications for a state-of-the-art neuromorphic hardware. Our results demonstrate an average 60\% lower crossbar requirement, 9x higher synapse utilization, 62\% lower wasted energy, and 3\% increase in model quality compared to an existing SNN mapping approach. In the future, we will explore the trade-offs involved in non-uniform tree decomposition of neural function.




\section*{Acknowledgment}
This work is supported by the National Science Foundation Award CCF-1937419 (RTML: Small: Design of System Software to Facilitate Real-Time Neuromorphic Computing).


\ifCLASSOPTIONcaptionsoff
  \newpage
\fi

\bibliographystyle{IEEEtran}
\bibliography{IEEEabrv,commands,disco,external}

\end{document}